\documentclass[11pt]{article}
\usepackage{acl-hlt2011}
\usepackage{algorithm2e}
\usepackage{amsmath}
\usepackage{times}
\usepackage{multirow}
\usepackage{latexsym}
\usepackage{graphicx}
\usepackage{url}
\usepackage{bm}
\usepackage{rotating}
\usepackage{caption}
\usepackage{subcaption}

\usepackage{commands}

\newcommand{\hidden}[1]{}

\title{Dating Texts without Explicit Temporal Cues}
\author{
Abhimanu Kumar \\
  School of Computer Science\\
  Carnegie Mellon University\\
  {\tt abhimank@cs.cmu.edu} \\\And
  Jason Baldridge \\
  Dept. of Linguistics\\
  University of Texas at Austin\\
  {\tt jbaldrid@mail.utexas.edu} \\\AND
  Matthew Lease \\
  School of Information\\
  University Of Texas at Austin\\
  {\tt ml@ischool.utexas.edu} \\
  \And
  Joydeep Ghosh \\
  Dept. of ECE\\
  University Of Texas at Austin\\
  {\tt ghosh@ece.utexas.edu} \\}

\date{}

\begin{document}

\maketitle

\begin{abstract}
% \abstractEn{
This paper tackles temporal resolution of documents, such as
determining when a document is about or when it was written, based
only on its text. We apply techniques from information retrieval that
predict dates via language models over a discretized timeline. Unlike most
previous works, we rely {\it solely}  on temporal cues implicit in the text. We
consider both document-likelihood and divergence based techniques and several smoothing 
methods for both of them. Our best model predicts the mid-point of
individuals' lives with a median of 22 and mean
error of 36 years for Wikipedia biographies from 3800 B.C. to the
present day. We also show that this approach works well when training
on such biographies and predicting dates both for non-biographical
Wikipedia pages about specific years (500 B.C. to 2010 A.D.) and for
publication dates of short stories (1798 to 2008). Together, our work shows
that, even in absence of temporal extraction resources, it is possible to
achieve remarkable temporal locality across a diverse set of texts.
% }

% \newpage
%This paper provides temporal resolution of documents, such as determining the
%mid-year of the life span of a person based on their Wikipedia biography or publication
%date of a story using its text. We propose and evaluate several models that
%select dates taking into accounting the likelihood of documents from different
%temporal periods. Our bayesian generative model provides the likelihoods of a test
%document belonging to different temporal periods. We compare this to a
%divergence based approach that constructs probability histograms using
%Kullback-Leibler Divergence between the language model for a test document and
%the language models for each period. We compare these two model with a third but
%similar approach approach used by Kumar et. al.~\shortcite{Kumar11-cikm}. Both
%our models perform considerable better on  all the three datasets that we use. We
%are able to predict the mid-life-span of individuals with a median error of 22
%and mean error of 36 years for Wikipedia biographies that have documents from
%3800 B.C. to present day. The median and mean errors for predicting the
%publication date for Guttenberg English short stories is 17 and 20 years
%respectively for a prediction range of more than 200 years.

\end{abstract}

\section{Introduction} \label{s:intro}

Temporal analysis of text has been an active area of research since the early
days of text mining with different focus in different disciplines. 
In early computational linguistics research it was primarily concerned
with the fine-grained ordering of temporal events
\cite{Allen:1983,Vilain:Beranek:1982}. Information retrieval research has
focused largely on time-sensitive document ranking~\cite{Dakka10,Li03cikm},
temporal organization of search results~\cite{AlonsoCIKM09},
and how queries and documents change over time~\cite{Kulkarni11wsdm}.

This paper explores temporal analysis models that use ideas present in
both computational linguistics and information retrieval. While some
prior research has focused on extracting explicit mentions of temporal
expressions~\cite{AlonsoCIKM09}, we investigate the feasibility of
using text alone to assign timestamps to documents. Following previous
document dating work~\cite{deJong05,Kanhabua08ecdl,Kumar11-cikm}, we
construct supervised language models that capture the temporal
distribution of words over {\it chronons}, which are contiguous atomic
time spans used to discretized the timeline.  Each chronon model is
smoothed by interpolation with the entire training set collection. For
each test document, a unigram language model is computed and used to
find the document's similarity with each chronon's language
model. This provides a ranking over chronons for the document,
representing the document's likelihood of being similar to the time
periods covered by each chronon~\cite{deJong05,Kanhabua08ecdl}. 

Our chronon models are learned from Wikipedia biographies spanning 3800 B.C. to 2010 A.D.
Wikipedia-based training is
advantageous since its recency enables us to control against stylistic vs.
content factors influencing vocabulary use (e.g. consider the difference between
William Mavor's 1796 discussion\footnote{\url{http://bit.ly/lKR8Aa}} of Sir
Walter Raleigh vs. a modern retrospective
biography\footnote{\url{http://en.wikipedia.org/wiki/Walter_Raleigh}}). This
contrasts with resources such as the Google n-grams corpus
\cite{GoogleN-Grams:2011}, which is based on publication dates, and thus
reflects information about when a document was written rather than what it is
about.

Our methods, all of which use the Wikipedia biographies for training models, are
evaluated on three tasks. The first is matched to the training data: predict the
mid-point of an individual's life based on the text in his or her Wikipedia
biography. Our best model achieves a median error of 22 years and a mean error
of 36 years. The second task is to predict the year for a set of events between
500 B.C. and 2010 A.D., using Wikipedia's pages for events in each
year.\footnote{\url{http://en.wikipedia.org/wiki/List_of_years}} The best model
gives a mean error of 36 years and median error of 21 years. The final task is
predicting the publication dates of short stories from the Gutenberg project
from the period 1798 to 2008.\footnote{\url{http://www.gutenberg.org}} In
comparison to biographies, these stories have far fewer mentions of historical
named entities that with peaked time signatures useful for prediction. This,
plus the difference in genre between Wikipedia biographies (training) and works
of fiction, stand to make this task more challenging. However, the distributions
learned from the biographies prove to be quite robust here: our best model
achieves a mean error of 20 years and a median error of 17 years from the true
publication date.

Our primary contribution is demonstrating the robustness and informativity of
the implicit temporal cues available in text alone, across a diverse set of
three prediction tasks.  We do so for document collections spanning hundreds and
thousands of years, whereas previous work has generally focused on relatively
short periods (decades) for recent time spans. Note that we use a robust
temporal expression identifier for English, Heidel-Time \cite{HeidelTime}, to
identify and remove dates from all texts for both training and testing. While
one could exploit a resource such as Heidel-Time to perform rule-based document
dating (possibly in combination with our methods and others such as \cite{chambers:doctimestamps}),
this work demonstrates that text-based techniques can be used effectively for languages for which 
such temporal extraction resources are not available (Heidel-Time has resources only for
English, German and Dutch). 

A second contribution is a thorough exploration of the information retrieval
approach for this task, including consideration of three different techniques
for smoothing chronon language models and a comparison of generative
(document-likelihood) and KL-divergence models for identifying the best chronon
for a test document. We find that straightforward Jelenik-Mercer smoothing
(basic linear interpolation) works the best, and that both document likelihood
and KL-divergence based approaches perform similarly. 

A specific task of interest in digital humanities is to identify and visualize
text sequences relating to the same time period across a collection of books. Our approach can
be used to timestamp subsequences of documents, which could be book-length
narratives or fictions, without explicit dates. 

%One of our primary contribution in this paper is to use the implicit cues
%present in text to assign timestamp. We propose 5 different models and evaluate them over 3
%different datasets for timestamp prediction. We train the models over a set of
%Wikipedia biographies and predict for the other 2 datasets using the same learnt
%model. We demonstrate that: a) the time related to document can be predicted
%without use of explicit dates present in the text (we discard all the explicit
%temporal cues present in the document on purpose), b) domain adaptation works
%quite well for temporal prediction tasks as our models trained on Wikipedia biographies perform
%equally well for Gutenberg short stories and event year texts, and c) the
%prediction models do not distinguish between the tasks to predict the time when
%a document was written and to predict the time of occurrence of the event
%that the text is about (the models perfom quite well for Gutenberg stories
%publication date prediction as well as time of year events from Wikipedia).

\section{Related Work} \label{s:related}

% Time sensitive model analysis have seen a spurt of activity in recent year in
% the field of IR and computational linguistics research. It has been an important
% area of focus for industry as well whether it is in terms of tracking news items
% over time or time based social network analysis.

{\bf Corpora for temporal evaluation.} With increased focus on temporal
analysis, there have been efforts to create richly annotated corpora to train
and evaluate temporal models, e.g. TimeBank \cite{timebank} and Wikiwars
\cite{Mazur:Dale:2010} %are two such annotated collection that 
were created to
provide a common set of corpora for evaluating time-sensitive  models. 
\shortcite{Loureiro11-GIS} use the above corpora to
resolve geographic and temporal references in text while
\shortcite{Chambers:Jurafsky:2008} use these to model event
structures.

{\bf Semantic based temporal models.} Time-sensitive models have also been
developed using semantic properties of data. \cite{Grishman:Huttunen:2002}
use semantic properties of web-data to create and automatically update a database
on infectious disease outbreaks. Other simpler approaches have been explored to
analyse literary and historical documents as well as recent datasets such as
tweets and search queries. Time based analysis of historical texts provides
important information as to how significant events happened in the past on a
temporal scale. The Google N-Grams
viewer\footnote{\url{http://ngrams.googlelabs.com/}}, which uses word counts
from millions of books and corresponding publication date, provides plots of
n-gram word sequences over a timeline \cite{GoogleN-Grams:2011}. This gives
useful insights into historical trends of events/topics and writing styles.
Time based analysis of tweets has gained popularity in recent years especially
to capture current trending topics for tracking news items and
market sentiment \cite{Zhang:Song:2010}. 

{\bf Time aware latent models.} Another approach for temporal
text analysis is latent variable based graphical models. Dynamic Topic Models
\cite{Blei:Lafferty:2006} are used to analyze the evolution of topics over time
in a large document collection \cite{Blei:Heckerman:2008}.
\shortcite{Wang:McCallum:2006} analyse variations in topic
occurrences over a large corpora for a fixed time period.
\shortcite{Manning:Hall:2008} investigate the history of ideas in
a research field though latent variable approaches. 
\shortcite{Chi:Zhu:2007} use graphical models for temporal analysis of
blogs and Zhang et al.\cite{Zhang:Song:2010} provide clustering techniques for
time varying text corpora through hierarchical Dirichlet processes for modeling time sensitivity.

{\bf Temporal analysis using conventional language models.} Time based
text analysis has been explored using conventional language model based
approaches for various applications e.g. time-sensitive query interpretation
\cite{Li03cikm,Dakka10}, time-based presentation of search
results~\cite{AlonsoCIKM09}, and modelling query and document changes over
time~\cite{Kulkarni11wsdm}. \cite{Li03cikm}, one of the early
temporal language models, use explicit document dates to estimate a more
informative document prior. More recently, \shortcite{Dakka10}
propose models for identifying important time intervals likely to be of interest
for a query incorporating document publication date into the ranking function.
\shortcite{AlonsoCIKM09} 
%provide motivation, overview and
%discussion of temporal analysis in information retrieval. They 
use explicit
temporal metadata and expressions as attributes to cluster documents and create timelines
for exploring search results.

Document dating---the task of this paper---is a closely related problem. 
~\shortcite{deJong05} follow a language model based approach to assign
dates to Dutch newspaper articles from 1999-2005 by partitioning the timeline
into discrete time periods.~\shortcite{Kanhabua08ecdl} extend
this work to incorporate temporal entropy and search statistics from Google
Zeitgeist. These approaches~\cite{deJong05,Kanhabua08ecdl} normalize the
evidence for each chronon by the whole collection. 
~\shortcite{chambers:doctimestamps} improve over these by including
linguistic constraints such as NERs, POS tagging and regular expression
based temporal relation constraints (e.g. ``after'', ``before'' etc.) and using
MaxEnt classifier for training. ~\shortcite{Kanhabua:Romano:2012} use
linguistics features such as sentence length, context, entity list in a document
etc. to discover events over twitter and assign time stamp by framing it as a
binary classification problem with the two classes as relevant and non-relevant. 
But, all these approaches worked for a small time range (6-10 years) but our datasets span 
around 5000 years and the evidence would die down after normalization. ~\shortcite{Kumar11-cikm} 
use divergence based methods and non-standard smoothing on Wikipedia biographies 
for the same task.  We perform our
experiments on two of their datasets, Wikipedia biographies and Gutenberg short
stories, and we compare their smoothing method with  standard Jelinek-Mercer and 
Dirichlet smoothing.

\section{Document Collections} \label{s:data}

Our models are trained and evaluated on three datasets \footnote{All three
will be released upon publication, including processing and extraction needed
for replication of experiments.}

\textbf{Wikipedia biographies (wiki-bio)}. The Wikipedia dump of English on
September 4, 2010 are used \footnote{\small
\url{http://download.wikimedia.org/enwiki/20100904/enwiki-20100904-pages-articles.xml.bz2}}
to obtain biographies of individuals who lived between the years 3800 B.C.\ to
2010 A.D.

We extract the lifetime of each individual via each article's \emph{Infobox}
\texttt{birth\_date} and \texttt{death\_date} fields. We exclude biographies
which do not specify one of the fields or which fall outside the year range
considered.  If the birth date is missing, we approximate it as 100 years before
the death date (similarly and conversely when the death date is missing). 
We perform this only to estimate the word distributions in the training set.
All such documents are discarded for validation and test. 
We treat the life span of each individual as the article's labeled time span. Note
that the distribution of biographies is quite skewed toward recent times, as
shown in  \fref{birth histogram}.
%, which plots the number of birth per year in
% the training set.

\begin{figure}
\begin{center}
% \iffigure
%   \centering 
  \includegraphics[width=0.5\textwidth]{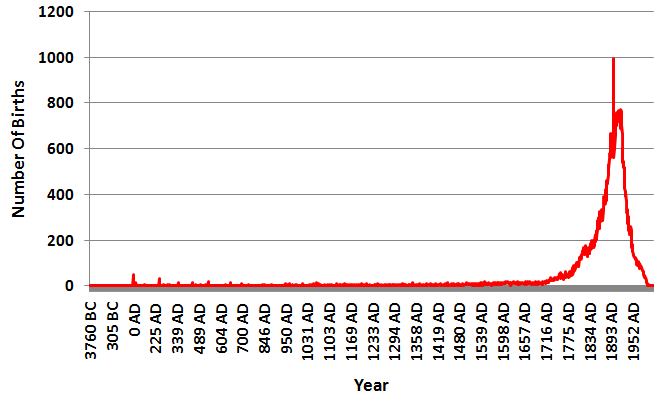}
% \fi
\caption{Graph of number of births per year in the Wikipedia biography training
set.}
\label{fig:birth histogram}
\end{center}
\end{figure}
% earlier 1956 and 1955
\begin{table}
\begin{center}
{\small
\begin{tabular}{c|p{5.8cm}}
Year & \multicolumn{1}{c}{Sample Text} \\ \hline
%  \multirow{5}{*}{1900 B.C.} & Port of Lothal is abandoned. \\
%  & Senwosret III (Twelfth Dynasty) started to rule.\\
%  & Proto-Greek invasions of Greece.\\
%  & Hittite empire in Hattusa, Anatolia.\\ 
%  & Fall of last Sumerian dynasty.\\ \hline 
%  \multirow{5}{*}{1000 B.C.} & Early Horizon period starts in the Andes.\\
%  & Iron is introduced in Ancient India.\\
%  & Phoenician alphabet is invented.\\
%  & Chavin culture starts in the Andes.\\
%  & Paracas culture starts in the Andes.\\ \hline
  \multirow{5}{*}{400 B.C.} 
 & The Carthaginians occupy Malta.\\
 & War breaks out between Sparta and Elis.\\
 & San Lorenzo Tenochtitlán is abandoned.\\ 
 & Thucydides, Greek historian dies.\\
 & The catapult is invented by Greek engineers.\\\hline
% \multirow{5}{*}{2 A.D.} & Cedeides becomes Archon of Athens.\\
%  & Birth of Deng Yu, Han Dynasty general and statesman.\\
%  & Publius Alfenus Varus and Publius Vinicius become Roman Consuls.\\
%  & Lucius Caesar, son of Marcus Vipsanius Agrippa and Julia the Elder, and heir
%  to the throne.\\ 
%  & The Chinese census shows nearly one million people living in Vietnam.\\
\multirow{5}{*}{50 A.D.} 
 & Claudius adopts Nero.\\
 & Phaedrus, Roman fabulist dies.\\
 & The Epistle to the Romans is written.\\
 & Abgarus of Edessa, king of Osroene dies.\\
 & Hero of Alexandria invents steam turbine.\\
 \hline 
 \multirow{5}{*}{1000 A.D.} 
%  & Middle Horizon period ends in the Andes.\\
 & Dhaka, Bangladesh, is founded.\\
 & The Diocese of Kołobrzeg is founded.\\
%  & Stephen I becomes King of Hungary, which is established as a Christian kingdom.\\
 & Garcia IV of Pamplona dies\\
 & Gunpowder is invented in China.\\ 
 & Middle Horizon period ends in the Andes.\\\hline
\multirow{5}{*}{2000 A.D.} 
%  & Stipe Mesic is elected president of Croatia.\\
 & Tate Modern Gallery opens in London.\\
 & Tuvalu joins the United Nations.\\
 & The last Mini is produced in Longbridge.\\
 & The Constitution of Finland rewritten.\\
 & Patrick O'Brian, English writer dies.\\
%  & Tuanku Syed Sirajuddin becomes Raja of Perlis.\\ 
 \hline
\end{tabular}
}
\end{center}
\caption{Sample text from 5 different years in wiki-year dataset.}
\label{tab:sample-wiki-years}
\end{table}

The resulting dataset contains a total of 280,867 Wikipedia biographies of individuals
whose lifetimes begin and end within the year range considered (3800
B.C. to 2010 A.D.). These biographies are randomly split into subsets
for training, development, and testing.  We remove documents from development and test
sets if either their \texttt{birth\_date} or \texttt{death\_date} missing. This
leaves us with 224,476 training articles, 8,358 development articles and 8,440
test articles.

\textbf{Wiki-year pages (wiki-year)}. Wikipedia has a collection of pages
corresponding to various years that describe the events that occurred for a given year.
\footnote{\url{http://en.wikipedia.org/wiki/List_of_years}}.
Each page has the corresponding year as its label and the text
contains all the events that occurred in that year -- some examples are shown in
\tref{sample-wiki-years}. Pages for years before 500 B.C. at times contain
events that span several years, so we restrict the documents to be those from
500 B.C. to 2010 A.D.\footnote{For example the events ``Proto-Greek invasions of
Greece.'', ``Minoan Old Palace (Protopalatial) period starts in Crete.'' etc.
are present in the text for 1878 as well as 1880 B.C. These occurred around 1880
B.C. but their exact occurrence date is unknown.} The 2,511 documents for this
span are divided into even years for development (1256 documents) and odd years
for testing (1255 documents).

Table~\ref{tab:sample-wiki-years} shows random sample lines from four wiki-year
pages. The lines are terse and the text as a whole contain very little temporal
expressions.

%\footnote{\url{http://www.gutenberg.org/ebooks/search.html/?default_prefix=subject_id&sort_order=downloads&query=94}}
\textbf{Gutenberg short stories (gutss)}. We collected 678
English short stories published between 1798 to 2008, obtained from the Gutenberg Project.
Whereas with Wikipedia biographies we use labeled time spans corresponding to
lifetimes, Gutenberg stories are labeled by publication year. The average,
minimum and maximum word count of these stories are (roughly) 14,000, 11,000 and
100,000 respectively. Stories are randomly split into a development and test set
of 333 and 345 documents, respectively.
% ML: we could center a time interval around the publication date: 
%Unlike the Wikipedia biographies, there is no equivalent span annotation for these stories. Instead, we use their publication date as an imperfect but useful target for temporal prediction. 

\textbf{Notation}. We refer to
biographies, stories and Wiki-Year pages alike as {\em documents}, and each
dataset as defining a document {\em collection} $c$ consisting of $N$ documents:
$c=d_{1:N}$.

\section{Model} \label{s:models}

Similar to previous work, we represent continuous time via
discrete units. Our formalization most closely follows that of Alonso et
 al.~\shortcite{AlonsoCIKM09}. The smallest temporal granularity we consider in
 this work is a single year. 

\subsection{Estimation} \label{s:represent}

% ML: notation indicates year y_e is excluded, otherwise we'd have (y_e+1)-y_s=\Delta
%, 

Let a {\em span} of multiple, contiguous years be some interval $\tau
= [y_s,y_e]$, where $y_s$ and $y_e$ refer to start and end years,
respectively. As noted in \S\ref{s:data}, we also know the year range covered by
each document collection and restrict our overall timeline correspondingly to
the span $\tau_o = [y_0,y_Y)$, covering a total of $y_Y-y_0=\Delta$ years.

A {\em chronon} is an atomic interval $x$ upon which a discrete timeline
is constructed \cite{AlonsoCIKM09}. In this paper, a chronon consists
of $\mathbf{\delta}$ years, where $\delta$ is a tunable parameter.
Given $\delta$, the timeline $T_\delta$ is decomposed into a sequence
of $n$ contiguous, non-overlapping chronons $\mathbf{x} = x_{1:n}$,
where $n=\frac{\Delta}{\delta}$.

% A document real or pseudo is represented by a single year. This is the mid-point
% of the chronon in case of a pseudo-
% 
% The mid-point, $\hat{y}$, of the chronon is the year of interest for our model.
% Given a test document, the model returns the mid-point of the most likely
% chronon as the prediction year.

A ``pseudo-document'' $d^x$ is created for each
chronon $x$ as the concatenation of all training documents whose labeled
span overlaps $x$. For example, for a chronon size $\delta$=25 years, the
biography of Abraham Lincoln (1809-1865) is included in pseudo-documents for each of the
chronons representing 1800-1825, 1826-1850, and 1851-1875. 

\label{s:smoothing}

A chronon model  $\Theta^x$ is estimated from the pseudo-document $d^x$ and
smoothed via interpolation with the collection. Chronon models are smoothed in
three ways: a) Jelinek-Mercer smoothing (JM) \cite{Zhai04}, b) Dirichlet
smoothing ~\cite{Zhai04}, and c) chronon-specific smoothing (CS)
\cite{Kumar11-cikm}. For all three, for each word $w$, $\hat{\Theta^d_w}$ can be
computed as a mixture of document $d$ and document collection $c$
maximum-likelihood (ML) estimates:
\begin{equation}
\hat{\Theta^d_w} = \lambda \frac{f^d_w}{|d|} + (1-\lambda) \frac{f^c_w}{|c|}~,
\end{equation}
where $f^d_w$ and $f^c_w$ denote the frequency of word $w$ in the document or
collection respectively, $|d|$ and $|c|$ are the document and collection
lengths, and the parameter $\lambda$ specifies the smoothing strength.  In 
case of Jelenik-Mercer smoothing, the value of $\lambda$ is chosen directly via
tuning over values from zero to one.

With Dirichlet smoothing, $\lambda$ is chosen as:
\begin{equation} 
\lambda=\frac{|d|}{|d| + \mu}
\end{equation}
$\mu$ is a hyper-parameter tuned on the development set.

Chronon-specific smoothing, in turn, is a special case of Dirichlet smoothing where:
\begin{equation}
\mu=\frac{\xi}{|V_{d^x}\cup V_{d_i}|}
\end{equation}
where $|V_{d^x}\cup V_{d}|$ denotes the
document-chronon specific vocabulary for some collection document $d_i$ and
pseudo-document $d^x$ and $\xi$ is a prior for hyper-parameter $\mu$ that is tuned on the development set.

\subsection{Estimation}

We calculate the affinity between each chronon $x$ and a document $d$ by
estimating the discrete distribution $P(x|d)$. In the next section, we use
$P(x|d)$ to infer affinity between $d$ and different chronons. The mid-point of
(see section~\ref{s:represent}) the most likely chronon is then returned as the
predicted year by the model. We define two primary models for estimating
$P(x|d)$. The first approach estimates the likelihood of $d$ for each chronon;
via Bayes rule, this is combined with a chronon prior to calculate the 
likelihood of each chronon for $d$. The second approach ranks chronons based on
the divergence between latent unigram distributions $P(w|d)$ and
$P(w|x)$~\cite{lafferty2001document}.

% \acomment{}{Do we describe the CIKM models? We definitely have to compare
% results}

% A second, more traditional approach identifies
% explicit year mentions in $d$ via regular expressions and estimates $P(x|d)$
% directly.  The derived model linearly interpolates the two primary models. 

% We begin by presenting relevant background on the language modeling approach.

% There are two ways to assign

% TODO:- partial and full counts
% 
% NOTE:- currently I have only results for ful-counts. (I do have results for
% partial counts but it is not the right on. I do partial count in different way
% than what Prof. Matt asked.)
%
\newcommand{\D}{\mathcal{D}}
\newcommand{\req}{\overset{rank}{=}}

\paragraph{Ranking by document likelihood}

The language modeling approach for information retrieval was originally
formulated as query-likelihood~\cite{Ponte98}. For our task, the document is the
``query'' for which we wish to rank chronons. We refer to this approach as
document-likelihood (DL).

We estimate $P(x|d) \propto P(d|x)P(x)$ via Bayes Rule. Assuming unigram
modeling, the likelihood of a test document $d$ is given by:

\begin{equation}
P(d|x) = \prod_{w \in d}\Theta^x_w
\end{equation}

\noindent
where the parameters of
$\Theta^x$ are estimated from the chronon $x$'s pseudo-document $d^x$, as described in
Section~\ref{s:smoothing}.

%and use a generative scheme as applied in the traditional language modeling(\cite{Zhai04}) to calculate $P(d|x)$ as:
%In the context of ranking documents for
%queries, prior work has shown that with a uniform prior, ranking by likelihood
%is equivalent to ranking by minimum KL-divergence. Consequently, the KL-ranking
%model in Equation (1) implicitly also covers the case of a uniform prior here.

Just as informed document priors (e.g. PageRank or document length) inform
traditional document ranking in information retrieval, an informed prior over
chronons has potential to benefit our task as well. We adopt a chronon prior
intuitively informed by the distribution of training documents over chronons:
\begin{equation}
P(x) =
\frac{|d_{train} \in d^x|}{\sum_{\forall y} |d_{train} \in d^y|}
\label{eqn:nonUniformPrior}
\end{equation}
where $d_{train}$ is a training document, $d^x$ is the pseudo-document for
chronon $x$ and $|d_{train} \in d^x|$ is the number of dated training documents
overlapping with chronon $x$.

\paragraph{Ranking by model comparison} \label{sec:kl}

Zhai and Lafferty \shortcite{Lafferty01b} propose ranking via KL-divergence
between a query and each collection document.  Kumar et
al.~\shortcite{Kumar11-cikm} use this approach to compute $P(x|d)$, which is
estimated by computing the inverse KL-divergence of $x$ and $d$ and normalizing
this value with the sum of inverse divergences with all chronons $x_{1:n}$:

\begin{equation} \label{e:inverse}
P(x|d) = \frac{\D(\Theta^{d} || \Theta^{x})^{-1}}{\sum_{y\in x_{1:n}} \D(\Theta^{d} ||
\Theta^{y})^{-1}} 
\end{equation} 

\noindent
It is straighforward to see that their formulation is rank equivalent to
standard model comparison ranking with negative
KL-divergence~\cite{deJong05,Kanhabua08ecdl}:

\begin{equation} \label{e:req}
P(x|d) 
\propto \D(\Theta^{d} || \Theta^{x})^{-1}
\req -\D_{KL}(\Theta^d ||\Theta^x)
\end{equation} 

Lafferty and Zhai showed such ranking is equivalent to generating the query
(i.e.\ query-likelihood) assuming a uniform document prior and the query model
being estimated by relative frequency ~\cite{Lafferty01b}. This means that for
our task, if we adopt a uniform prior over chronons and estimate the document
model by relative frequency, then KL-ranking and document-likelihood approaches
will be rank equivalent.

\label{s:inference}

\paragraph{Prediction} Having determined $P(x|d)$, we choose the midpoint
$\hat{y}$ of the most likely chronon; for a chronon $x=[y, y+\delta]$, the
mid-point is $\hat{y} = y+\delta/2$.

%==============================================================================
%==============================================================================
\section{Experimental Setup} \label{s:evalsetup}
\label{sec:metrics}

%\subsection{Data}

\paragraph{Data.} To test the ability of word-based models to predict timestamps
for documents, all temporal expressions identified in each document using the
Heidel-Time temporal tagger~\cite{HeidelTime} are removed. All numeric tokens
and standard stopwords are also removed. The remaining tokens produce a
vocabulary of 374,973 words for the entire Wikipedia biography corpus.
Heidel-Time also provides the first two dates present in the text, which we use
as a strong baseline for the biography task.

\paragraph{Tuning and smoothing} For each model+task, we tune the parameters
$\delta$, $\mu$, $\xi$, and $\lambda$ over the development sets of the
corresponding dataset. As in prior work~\cite{deJong05,Kanhabua08ecdl,Kumar11-cikm}, we smooth chronon
pseudo-document language models (for all models as well as smoothing techniques)
but not document models. While smoothing both may potentially help, smoothing
the former is strictly necessary for KL-divergence to prevent division by zero.

\paragraph{Target predictions} For Wikipedia biographies, the predicted $\hat{y}$
represents the mid-point of the individual's life span; for wiki-years, 
it is the year of the events on the page, and for Gutenberg short stories it
is the publication date of the story. In later sections we will present the baseline
predictions for $\hat{y}$ for each dataset.

\paragraph{Error Measurement} When predicting a single year for a document, a natural
error measure between the predicted year $\hat{y}$ (mid-point) and the actual
year $y^*$ is the difference $|\hat{y}-y^*|$. We compute this difference for
each document, then compute and report the mean $\bar{y}$ and median {\bf
$\tilde{y}$} of differences across documents. Similar distance error measures
have also been used with document geolocation
\cite{eisenstein10,wing-baldridge:2011:ACL}.

\paragraph{Baselines} For Wikipedia biographies the first baseline
(\textbf{baseline-ht}) is the mid-point of the first two temporal-dates extracted
by Heidel-Time~\cite{HeidelTime}. This is a highly effective baseline since 
it is often the case in Wikipedia biographies that the first
two dates are the birth and death dates. The second
baseline  for biographies is to always predict the year that has
greatest number of biographies spanning it, which is 1915 (\textbf{baseline-1915}). 
For Gutenberg stories, we take 1903, the
midpoint of the range of publication dates (1798-2008) as the baseline 
(\textbf{baseline-1903}). For wiki-years, the baseline is the midpoint of the
prediction range i.e. $\frac{-500+2010}{2} = 755$ (\textbf{baseline-755}). 
This assumes that one knows a rough range of possible publication dates, which
is reasonable for many applications and thus provides a good reference for comparison.

We also report \textbf{oracle error} which is the
mean and median error which would occur if a model always picked the correct 
chronon. This error arises because chronons span multiple years; large chronons
in particular will have higher oracle error (but may perform better for actual
prediction due to better model estimation).

% \begin{figure*}
% \hspace*{-2mm}
% \begin{minipage}[b]{3.2in}
% % \epsfxsize=.9\linewidth
% % \iffigure
% %   \rotatebox{0}{
%   \includegraphics[width=8.0cm]{graphs/TinyGraphsDelta.png}%}
% % \fi
% \end{minipage}
% \hspace*{.1cm}
% \begin{minipage}[b]{3.6in}
% % \epsfxsize=.9\linewidth
% % \iffigure
% %   \rotatebox{0}{
%   \includegraphics[width=8.5cm]{graphs/TinyGraphsSmoothing.png}%}
% % \fi
% \end{minipage}
% \hspace*{5cm}(a)\hspace*{7.5cm}(b)
% \caption{Tuning for $\delta$ (fig. a) and smoothing parameters (fig. b) over
% wiki-bio and wiki-years datasets for KL model. $\xi$ (for CS) and $\lambda$ (for
% JM) are fixed at 0.01 and 0.99 respectively in fig a (for $\delta$ tuning).}
% \label{fig:tuning-wiki-bio-years}
% \end{figure*}

% \begin{figure}
% \begin{center}
% \begin{subfigure}[b]{0.4\textwidth}
%                 \centering
%                 \includegraphics[width=\textwidth]{graphs/TinyGraphsDelta.png}
%                  \caption{}
%                 \label{fig:a}
% \end{subfigure}
% % \vspace*{1cm}
% \begin{subfigure}[b]{0.4\textwidth}
%                 \centering
%                 \includegraphics[width=\textwidth]{graphs/TinyGraphsSmoothing.png}
%                  \caption{}
%                 \label{fig:b}
% \end{subfigure}
% % \hspace*{5cm}(a)\hspace*{7.5cm}(b)
% \caption{Tuning for $\delta$ (fig. a) and smoothing parameters (fig. b) over
% wiki-bio and wiki-years datasets for KL model. $\xi$ (for CS) and $\lambda$ (for
% JM) are fixed at 0.01 and 0.99 respectively in fig a (for $\delta$ tuning).}
% \label{fig:tuning-wiki-bio-years}
% \end{center}
% \end{figure}

\begin{figure}[h!]
  \centering
    \includegraphics[width=0.5\textwidth]{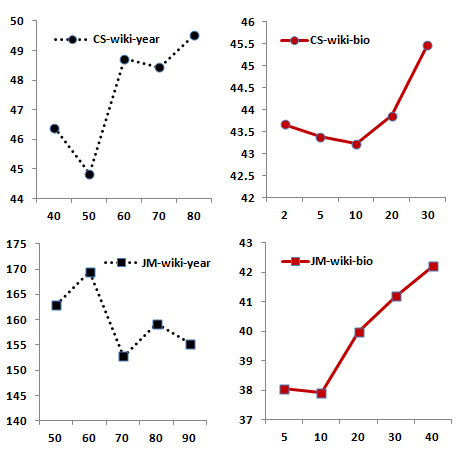}
    \caption{Tuning for $\delta$ over wiki-bio and wiki-years datasets for KL
    model. $\xi$ (for CS) and $\lambda$ (for JM) are fixed at 0.01 and 0.99
    respectively.}
    \label{fig:tuning-wiki-bio-years-delta}
\end{figure}

\begin{figure}[h!]
\centering
\includegraphics[width=0.5\textwidth]{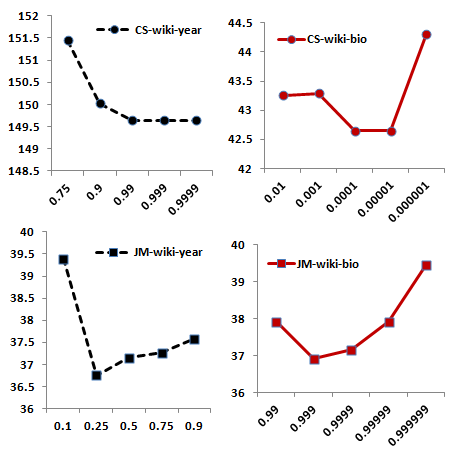}
\caption{}
\label{fig:b}
% \hspace*{5cm}(a)\hspace*{7.5cm}(b)
\caption{Tuning for smoothing parameters ($\xi$ and $\lambda$) over
wiki-bio and wiki-years datasets for KL model.}
\label{fig:tuning-wiki-bio-years-para}
\end{figure}

%==============================================================================
%==============================================================================
\section{Results}
\label{sec:exps}

% \newcolumntype{P}[1]{>{\raggedright}p{#1}}

\subsection{Parameter tuning}

We begin with year prediction experiments on the development sets to
tune the parameters $\delta$, $\xi$ or $\mu$. We parametrize $\mu$ as a
function of the average chronon size in the training set:

\begin{equation}
\mu = \lceil\rho\bar{c}\rceil
\end{equation}

\noindent
% where $\bar{c}$ is the average chronon size in the training set. 
$\bar{c}$ is a constant whose value is dependent upon the model and the task.
The value of $\rho$ is tuned over the validation set.
 
\paragraph{Choice of chronon size and smoothing parameters.}  We tune the chronon
size ($\delta$) over the validation set and tune the smoothing parameters
$\lambda$, $\rho$, and $\xi$ (depending on the type of smoothing) for the
best $\delta$ obtained. For $\delta$ tuning we assign an arbitrary value to the
smoothing parameter $\lambda$. The $\delta$ is tuned for each dataset and KL
model with CS and JM smoothings. DL model with Dirichlet/JM smoothing and
KL model with Dirichlet smoothing use the same best $\delta$ obtained for KL model
with JM smoothing on the respective datasets. For each dataset, model and smoothing
triad, the smoothing parameter $\lambda$, $\xi$, or $\rho$ is tuned. Tuning is 
performed to minimize the mean error on the development sets. The search space 
for smoothing parameters $\xi$, $\lambda$
and $\rho$ includes \{ $1e-12$, $1e-11$, \ldots, 0.1, 0.25, 0.75, 0.9, 0.99, \ldots, 0.999999999 \}

Figures~\ref{fig:tuning-wiki-bio-years-delta} and~\ref{fig:tuning-wiki-bio-years-para} shows the tuning of
$\delta$ and smoothing parameters ($\lambda$ for JM and $\xi$ for CS) for the wiki-bio and
wiki-years dataset. All triplets formed by KL/DL model $\times$ JM/Dirichlet
smoothing $\times$ wiki-years/wiki-bio/gutss dataset use the optimum
chronon-size obtained for the respective datasets from the KL model with JM smoothing.

%Table~\ref{tab:tuning-results} provides the best results for each triplet (model,
%data, smoothing) and the baseline for each dataset using the above described
%tuning scheme on the validation set. KL model with JM smoothing stands out to be
%the best across all the there datasets over the validation set. 

From Figure~\ref{fig:tuning-wiki-bio-years-para} the mean error curve is generally
smooth for $\lambda$ and $\xi$ unlike the $\delta$, chronon-size parameter
(figure~\ref{fig:tuning-wiki-bio-years-delta}). This makes smoothing the LMs robust to a range of values.
The $\delta$ has more fluctuation even in the optimal neighborhood, which makes tuning chronon-size
more critical. A straightforward strategy to reduce this sensitivity is to 
smooth chronon models based on the word distributions of neighboring chronons 
as well as interpolating with the collection model, which we intend to explore 
in future. The optimal chronon sizes for the three datasets are 10 years for 
wiki-bio and gutss and 50 years for wiki-year.

\subsection{Test results.}

% \begin{table} [!htb]
% \begin{center}
% \scalebox{0.9}{
% \begin{tabular}{l|l|c||c|c}
% Dataset & Model & Smoothing & $\bar{y}$ & $\tilde{y}$
% \\
% \hline
% baseline & 	  &   &  306.64  &  0.0 \\	%8440 test docs
% wiki-bio & KL & $\xi$ &  42.79  &  22.50 \\
% wiki-bio & KL & $\lambda$ &  37.40  &  22.50  \\
% wiki-bio & KL & $\rho$ &  38.09  & 22.00  \\
% wiki-bio & Bayesian & $\lambda$ &  37.35  &  22.50  \\
% wiki-bio & Bayesian & $\rho$ &  38.08  &  22.00 \\
% \hline
% baseline & 	  &			 & 37 & 50 \\
% gutss & KL & $\xi$ & 39.61  & 19.00  \\
% gutss & KL & $\lambda$ &  22.89  & 19.00  \\ %0.9999999
% gutss & KL & $\rho$ & 37.33   & 22.00  \\
% gutss & Bayesian & $\lambda$ & 22.92   & 19.00 \\ %0.9999999 %
% gutss & Bayesian & $\rho$ &  37.39  &  23.00 \\
% \hline
% baseline & 	  &   & 978 & 489 \\
% wiki-year & KL & $\xi$ & 143.57  & 30.00  \\
% wiki-year & KL & $\lambda$ &  37.92  & 20.00  \\
% wiki-year & KL & $\rho$ & 60.65   & 22.00 \\
% wiki-year & Bayesian & $\lambda$ &  37.92  & 20.00 \\
% wiki-year & Bayesian & $\rho$ &  52.08 &  20.00 
% 
% \end{tabular}
% }
% \end{center}
% \caption{Model Results on test set. JM=Jelinek-Mercer and CS=chronon-specific,
% BM=Bayes Model, and non-U=non-uniform prior}
% \label{tab:test-results}
% \end{table}

Table~\ref{tab:test-results} shows the results for the various models on the
test sets for all three datasets, using the parameters tuned on the
corresponding development sets.

\paragraph{Wiki-bio} 
The models beat both baselines easily. Note that baseline-ht is quite
strong for a large number of documents: it gives a median error of zero since
over half of the documents have birth and death dates as their first dates.
Nonetheless, it fails entirely for many documents, and obviously has limited
applicability. The models all reduce error by one half in comparison to
baseline-1915. The best model (DL + JM smoothing) achieves a mean error of 37.4
years, which is quite strong given that the prediction range is 5810 years.  The
mean oracle error for the best model is 2.5 years. The mean and median error was
36.6 and 22.0 years for the best performing model (DL + JM smoothing) on the
development set.

\paragraph{Wiki-years}
The models beat baseline-755 comfortably. Despite the fact that the documents 
are relatively short and that any given document contains a number of often 
unrelated events (and thus low counts per word type), the results are in line 
with those for wiki-bio, with mean error of 37.9 and median error of 20 years 
for the best models. The mean oracle error, 12.4 years, for this dataset is 
higher due to the larger chronon size. The KL model with JM smoothing provided
the best mean and median error of 36.7 and 21.0 years respectively over
development set.

% Using this and
% figure~\ref{fig:wiki-years-meanErrorHistogram} we can say that the model is wrong by more than 56 years for majority of the documents in B.C.
% range.

\paragraph{Gutss} All models except the one that uses chronon-specific smoothing 
with KL-divergence outperform baseline-1905 on mean error, and even that one is
better on median. Since these are works of fiction with few historical entities 
mentioned, the mean error of 22.9 and median error of 19.0 of the best models 
indicate that the approach is quite capable of exploiting implicit temporal cues 
of basic vocabulary choices. Also, recall that the model is trained on Wikipedia; 
this demonstrates that this choice of training set works well as the basis for 
predictions on other domains. The mean oracle error (for chronons of 10 years) 
is 2.5 years. For the development set, the mean and median error was 20.4 and 
17.0 years for the best performing model (DL + JM smoothing).

\begin{table}[h]
\begin{center}
\scalebox{1.0}{
\begin{tabular}{l|l||c|c}
Data & Model(Smooth.) & $\bar{y}$ & $\tilde{y}$
\\
\hline
\multirow{7}{1mm}{\begin{sideways}\parbox{15mm}{wiki-bio}\end{sideways}} 
& baseline-ht 	&  306.6  &  0.0 \\	%8440 test docs
& baseline-1915   &  81.1  &  38.5 \\	%8440 test docs
& KL($\xi$=$10^{-4}$) &  42.8  &  22.5 \\
& KL($\lambda$=0.999) &\bf  37.4  &  22.5  \\
& KL($\rho$=$10^{-6}$) &  38.1  & \bf 22.0  \\
& DL($\lambda$=0.999) & \textbf{\textit{37.4}}  &  \textit{22.5} \\
& DL($\rho$=$10^{-6}$) &  38.0  &  \bf 22.0 \\
\hline
\multirow{6}{1mm}{\begin{sideways}\parbox{15mm}{wiki-year}\end{sideways}} 
& baseline-755 	    & 627 & 627 \\%& 978 & 489 \\
& KL($\xi$=0.99) & 143.6  & 30.0  \\
& KL($\lambda$=0.25) &  \textbf{\textit{37.9}}  & \textbf{\textit{20.0}}  \\
& KL($\rho$=0.01) & 60.6   & 22.0 \\
& DL($\lambda$=0.50) & \bf 37.9  & \bf 20.0 \\
& DL($\rho$=0.01) &  52.1 &  \bf 20.0 \\
\hline
\multirow{6}{1mm}{\begin{sideways}\parbox{15mm}{gutss}\end{sideways}} 
& baseline-1905 	 		 & 37 & 50 \\
& KL($\xi$=$10^{-3}$) & 39.6  & \bf 19.0  \\
& KL($\lambda$=0.999) & \bf 22.9  & \bf 19.0  \\ 
& KL($\rho$=$10^{-6}$) & 37.3   & 22.0  \\
& DL($\lambda$=0.999) & \textbf{\textit{22.9}}   & \textbf{\textit{19.0}} \\ 
& DL($\rho$=$10^{-6}$) &  37.4  &  23.0 
\end{tabular}
}
\end{center}
\caption{Test set results. $\lambda$=JM, $\xi$=CS, and 
$\rho$=Dirichlet smoothing. DL uses the non-uniform chronon prior. The best
results are bolded, and the results of the best model on the corresponding
development set are italicized.}
\label{tab:test-results}
\end{table}

%The standard smoothing techniques (JM and Dirichlet) perform better than the
%Kumar et al's CS smoothing accross models and datasets. Among the conventional
%smoothings, JM performs slightly better than Dirichlet given the dataset and
%model. From Table~\ref{tab:test-results} the median error for wiki-bio baseline
%is zero which is not surprising. For Wikipedia biographies, the first two dates
%present are birth and death dates in order with very high probability. And once
%the baseline picks it up its error goes to zero. This happens for atleast more
%than half of the documents being predicted which results in the median error as
%zero.

\subsection{Output analysis}

Using the output on the development set, we find interesting patterns in the
predictions made by the models and the way they use the words as evidence.

\paragraph{Time warps} 

\shortcite{flickRwormholes:2010} used geotags on Flickr 
images to identify wormholes---locations that are not physically near but 
which are nonetheless similar to one another. We observe some similar 
patterns, in our case {\it time warps}, in our dataset.
These are particularly prominent in wiki-year documents due to their 
terseness as these are list of events that happened in a given year. 
Besides the models trained on wiki-bio
set add to this phenomenon as the context for the two datasets are
slightly different. A cluster of dev event years from between 250 to
150 A.D. (e.g. wiki-years 234, 214, 152, 156 etc.) are predicted to be
in 2nd century B.C. (200 B.C. to 150 B.C.) by our model.  These event
years are very short with an average length of 40-50 words per
document. The discriminatory tokens present in these texts include: {\it
  Roman, Empire, Kingdom, Han, Dynasty, China, Selucid, Greek,
  etc.}. In the 200-150 B.C. period, all the documents in
training set are about Greek/Selucid, Roman and Chinese (mostly from
Han dynasty) emperors/personalities (e.g. Attalus I, Eratosthenes,
Plautus, Emperor Gaozu of Han, Emperor Hui of Han, Zhang Qian, Emperor
Wen of Han etc.) and contain similar prominent terms as the wiki-year
event texts. This common collection of terms pushes the model to resolve 
wiki-year texts to 2nd century B.C. This happens because of the relative 
frequency of such terms in B.C. and A.D.: although these terms are 
present in the A.D. chronons, their
proportion with respect to other terms is much smaller. Test documents 
that contain these terms are thus attracted to the B.C. chronons since 
they have these terms in generally higher proportion.

\begin{table*}
\begin{center}
\begin{tabular}{c|c|c|c|c}
\hline
meriwether & komatsu & capote & cranmer & payload \\ \hline 
morelos & kido & stopes & sap & laila \\ \hline 
hem & shakuntala & anthrax & scooby & crayon \\ \hline 
plutarch & sampaguita & woodbury & untimely & teleplay \\ \hline 
tele & electorates & derivatives & polygram & wavelength \\ \hline
%note that the for the least predictive they are mostly uncommon present in
% atmost 2 dev documents
\end{tabular}
\end{center}
\caption{ Top 25 most predicitve words in descending
order (left to right and top to bottom) from wiki-bio dev set. }
\label{tab:top20-words}
\end{table*}

\begin{table*}
\begin{center}
\begin{tabular}{c|c|c|c|c}
\hline
oneself & primari & ssu & thebes & porphyry \\ \hline 
lysias & confucius & morality & romana & matteo \\ \hline 
unbroken & goodness & timpul & tarii & grout \\ \hline 
sinop & cynical & tub & crates & lantern \\ \hline 
bite & phila & transaction & corporeal& conciliation \\ \hline

\end{tabular}
\end{center}
\caption{ Bottom 25 least predicitve words in descending
order (left to right and top to bottom) from wiki-bio dev set. }
\label{tab:bottom20-words}
\end{table*}

Another interesting cluster is short documents containing
similar terms from 200-800 A.D. that are resolved to the mid-6th century A.D. The short
wiki-year texts (e.g. years 246, 486, 750, 822, etc.) contain co-occurring
set of terms like {\it Byzantine, Empire, Roman, Arab, Conquest, Islam}, and {\it Caliphate}. These
short year events text contain events
related to mostly Byzantine wars, emperors, Islamic/Arab conquest, Caliphates
etc. These are resolved to the mid-6th century A.D. period that predominantly contains
biographies of Islamic Caliphates (e.g. Abd al-Malik, Abu Bakr, Ali, Umar etc.)
and Byzantine emperors and prominent personalities (e.g. Maurice, Fausta,
Constans II etc.) which has predominant terms such as: {\it Byzantine, Empire,
Caliph, Islam}, and {\it conquest}.

\paragraph{Discriminative Words }
Table~\ref{tab:top20-words} and~\ref{tab:bottom20-words} shows the top and
bottom 25 words in the descending order of their strengths, where the predictive strength score
of a word $w$ is calculated as average prediction error of all the documents that contain the
word $w$. The majority of the words that are most predictive are uncommon nouns, especially
uncommon last names or famous titles e.g. {\it capote}, {\it komatsu}, and {\it cranmer}.
Words such as {\it tele}, {\it wavelength}, {\it electorates},
{\it teleplay}, {\it sap} (the company) also have strong temporal connection as
these have never been used before 19th century. The
least predictive ones are mostly common words such as {\it
goodness}, {\it oneself}, {\it morality}, {\it tub}, {\it crates}, and {\it lantern}. The 
uncommon words among the least predictive are generally present in just one
or two documents for which our model performs very poorly. It is highly likely that these words
might be inducing those warps due to their predominance and uniqueness. 
% This again points 
% to the likely utility of smoothing chronons based on their neighboring chronons
% as well as the collection.

\section{Conclusion} \label{s:conclusion}

Using words alone, it is possible to identify the time period that a document is
about (via the Wikipedia datasets) or the time period in which it was written
(via the Gutenberg dataset). In the former case, the presence of named entities
dominates the texts, and their names provide strong evidence for particular
historical periods. For the latter, the texts are fictional (including science
fiction), and they rarely mention historical entities. For these, general terms
that are indicative of a given time period dominate the prediction.
Interestingly, the models that are used (successfully) for this later task are
trained on Wikipedia biographies about historical individuals, but which were
written in the last decade.

The predictions made by our models provide a natural counterpart to other
temporally sensitive models of word choice, such as Dynamic Topic Models (DTMs)
~\cite{Blei:Lafferty:2006}. DTMs assume that documents are labeled with dates;
our model could thus be used to create labels for an otherwise un-dated set of
documents which can then be analyzed with DTMs. An important aspect of our work is that it opens
opportunities for analyzing sub-parts of documents, such as chapters, sections
and paragraphs of books. Consider, for example, Samuel Goodrich's ``The Second Book of History''
from 1840, which covers thousands of years of history for many parts of the
world.

Of course, many texts include explicit dates, and exploiting their presence via
approaches such as~\shortcite{chambers:doctimestamps} would only strengthen our predictions. Also,
they create opportunities for using weaker, but more pin-pointed, supervision:  strings identified as dates with
high-confidence can be pivots for learning word distributions. This would
obviate the need for labeled training material such as Wikipedia biographies,
and thereby enable our methods to be used and adapted for a wide variety of
genres. Given decent temporal expression identifiers for other languages, this
could be used to bootstrap models for more languages as well.

\section*{Acknowledgments}
% We thank the anonymous reviewers for their valuable feedback and
% suggestions. 
This work was partially supported by a grant from the Morris
Memorial Trust Fund of the New York Community Trust and a Temple Fellowship.

\bibliographystyle{acl}
\bibliography{mybib}

\end{document}